\documentclass[a4paper]{article}

\usepackage{INTERSPEECH2021}
\usepackage{appendix}
\usepackage{subcaption}
\usepackage{multirow}
\usepackage{tabularx}
\usepackage[algo2e]{algorithm2e}
\usepackage{graphicx}
\usepackage{hyperref}
\usepackage{microtype}
\usepackage[T1]{fontenc}
\usepackage{placeins}
\usepackage{latexsym}
\usepackage{graphicx}
\graphicspath{{figures/}}

\usepackage{subcaption}
\usepackage{booktabs}
\usepackage{makecell}
\usepackage{float}
\graphicspath{{figures/}}

\usepackage{mathtools}
\usepackage{amssymb} % widetilde
\usepackage{mathrsfs} % mathscr
\usepackage{bbm}  % for \mathbbm 1
\usepackage{algorithm}
\usepackage{algpseudocode}

\algnewcommand{\IIf}[1]{\State\algorithmicif\ #1\ \algorithmicthen}
\algnewcommand{\EndIIf}{\unskip\ \algorithmicend\ \algorithmicif}

\algnewcommand{\LineComment}[1]{\State \(\triangleright\) #1}
\algnewcommand{\IfThenElse}[3]{% \IfThenElse{<if>}{<then>}{<else>}
  \State \algorithmicif\ #1\ \algorithmicthen\ #2\ \algorithmicelse\ #3}
  
 \algnewcommand{\IfThen}[2]{% \IfThenElse{<if>}{<then>}{<else>}
  \State \algorithmicif\ #1\ \algorithmicthen\ #2}

\usepackage{nameref}  % reference sections by name

\usepackage{xspace}  % for onedot
\usepackage{enumitem}

\newcommand{\x}{\boldsymbol x}

\makeatletter
\DeclareRobustCommand\onedot{\futurelet\@let@token\@onedot}
\def\@onedot{\ifx\@let@token.\else.\null\fi\xspace}

\makeatother

\usepackage{tikz, pgfplots}
\usepackage{pgffor, pgfmath}

\pgfplotsset{
    compat=1.15,
    grid style={darkgray},
    minor grid style={gray!20},
    major grid style={gray!20},
    axis line style = { darkgray }, 
    every axis plot/.append style={line width=1.5pt, mark options=solid, mark size=4pt},
    legend style={draw = darkgray, rounded corners=0pt, fill = white, font=\Large},
    tick style ={color = gray!30 },
    tick label style={font=\normalsize},
    label style={font=\normalsize},
}

\title{Impact of Encoding and Segmentation Strategies \\
on End-to-End Simultaneous Speech Translation}
\name{Ha Nguyen$^{1,2}$, 
    Yannick Est{\`e}ve$^2$, 
    Laurent Besacier$^{1,3}$}
\address{
   $^1$LIG - Universit{\'e} Grenoble Alpes, France  \\
    $^2$LIA - Avignon Universit{\'e}, France \\
    $^3$Naver Labs Europe, France}
\email{manh-ha.nguyen@univ-grenoble-alpes.fr, yannick.esteve@univ-avignon.fr, laurent.besacier@naverlabs.com}

\begin{document}

\maketitle
\begin{abstract}
  
Boosted by the simultaneous translation shared task at IWSLT 2020, promising end-to-end online speech translation approaches were recently proposed. They consist in incrementally encoding a speech input (in a source language) and decoding the corresponding text (in a target language) with the best possible trade-off between latency and translation quality.
This paper investigates two key aspects of end-to-end simultaneous speech translation: (a) how to encode efficiently the continuous speech flow, and (b) how to segment the speech flow in order to alternate optimally between reading (R: encoding input) and writing (W: decoding output) operations.
We extend our previously proposed end-to-end online %model
decoding strategy and show that while replacing BLSTM by ULSTM encoding degrades performance in offline mode, it actually improves both efficiency and performance in online mode. We also measure the impact of different methods to segment the speech signal (using fixed interval boundaries, oracle word boundaries or randomly set boundaries) and show that our best end-to-end online %model 
decoding strategy is surprisingly the one that alternates R/W operations on fixed size blocks on our English-German speech translation setup.
%not influenced by segmentation. For instance, alternating R/W operations on fixed size blocks gives the best quality/latency trade-off on our English-German speech translation setup.

\end{abstract}

\vspace{-5pt}
\section{Introduction}

\textit{Online} (also known as \textit{simultaneous}) machine translation refers to automatic translation systems which start generating an output hypothesis before the entire input sequence has been consumed \cite{bangalore2012real,sridhar2013segmentation}. Emerging recently as a challenging task, it has been witnessing several works proposed in text-to-text ($T2T$) translation \cite{Ma19acl, Arivazhagan19acl, Ma20iclr,elbayad:hal-02962195}, and in speech-to-text ($S2T$) translation \cite{DBLP:journals/corr/abs-1808-00491,elbayad:hal-02895893,han-etal-2020-end, nguyen2021empirical_icassp}, which attempt to deal with the low latency constraint imposed by the task. Following the \textit{wait-k} policy originally proposed for $T2T$ \cite{Ma19acl} and proven effective when applied to $S2T$ \cite{elbayad:hal-02895893,han-etal-2020-end}, our previous work \cite{nguyen2021empirical_icassp} introduced an adaptive version of \textit{wait-k} which leverages any pre-trained end-to-end offline speech translation  model for online speech translation.
%and allows the system to write several several output tokens at a time. 
However, the model proposed in \cite{nguyen2021empirical_icassp} had a speech encoder based on a \textit{Bi-directional} Long Short-Term Memory (BLSTM) \cite{LSTM}  which was not efficient in online mode since re-encoding of the full input was needed each time a new speech block was read.
%online processing  , and operating on fixed segmentation blocks. 

We show in this work that while replacing BLSTM by \textit{Uni-directional} Long Short-Term Memory (ULSTM) encoding degrades performance in offline mode, it actually improves both efficiency and performance in online mode (this observation was also made for online $T2T$ translation by \cite{elbayad:hal-02962195}). We also investigate how to segment the speech flow in order to alternate optimally between reading (R: encoding input) and writing (W: decoding output) operations. The contributions of this work are the following:

%BLSTM speech encoder while does not show superiority in terms of BLEU/AL trade-off over its \textit{Uni-directional} LSTM (ULSTM) counterpart, significantly underperforms ULSTM in terms of decoding speed. 

\begin{itemize}
    \item Showing that ULSTM speech encoder when using the same (\textit{re-encode}) encoding strategy yields better inference speed and performance in comparison with BLSTM speech encoder, 
    \item Further improving inference speed and performance of ULSTM speech encoder using a new encoding strategy (ULSTM \textit{Overlap-and-Compensate}),
    \item Analyzing the impact of speech flow segmentation on the BLEU/Latency trade-off, comparing three segmentation methods: fixed interval boundaries, oracle word boundaries or randomly set boundaries.
    %\item Introduce a measure to measure online translation complexity of utterances.
    %Analyzing the impact of the sentence complexity on the model's behaviors for the low latency task.
\end{itemize}
\vspace{-10pt}
\section{Background on low latency neural speech translation}
%\vspace{-5pt}
\subsection{Decoding strategies}
Real-life applications require translation systems to start emitting output translation partially before the input sequence is made fully available. Such a low latency constraint has been imposing great challenge to neural sequence-to-sequence translation models, despite their state-of-the-art performance on offline translation tasks. Notable efforts have been going into optimizing quality/latency trade-off of the neural online translation systems, including \cite{cho2016can} who introduces a waiting policy which alternates READ/WRITE operations. Inspired by \cite{cho2016can}, \cite{dalvi2018incremental} designs a static read and write decoding policy, which first reads $S$ input tokens, and alternates between a same number of WRITE and READ operations until the entire source sequence is consumed. In the same spirit, \cite{Ma19acl} proposes a \textit{wait-k} decoding policy which reads $k$ source tokens at the first step, and then alternates single WRITE/READ operations. 

Several works on online automatic speech translation got decent results when adapting \textit{wait-k} policy to their task, including \cite{DBLP:journals/corr/abs-1808-00491,elbayad:hal-02895893,han-etal-2020-end, nguyen2021empirical_icassp}. \cite{han-etal-2020-end} made an attempt to build an end-to-end online system which first reads $k$ input frames, then alternates between writing one output token or reading the next $s$ input frames. \cite{nguyen2021empirical_icassp} extends this work, modifying their decoding policy to be able to emit more than one (and maximum $N$) output tokens at a time. 
The policy of \cite{nguyen2021empirical_icassp} allows them to exploit any pre-trained offline model in an online decoding mode. However, they only experiment with pre-trained models whose speech encoders use BLSTM layers. \cite{elbayad:hal-02962195} shows that, in online mode, BLSTM models might be an unnecessarily costly choice, and therefore advocates for using ULSTM models instead (for text translation). 
In this work, we explore the use of ULSTM models, and make a comparison with their BLSTM counterpart for low latency end-to-end speech translation. We also experiment alternative speech segmentation policies to \cite{nguyen2021empirical_icassp}.
\vspace{-5pt}
\subsection{Evaluation metrics}
Performance of online translation systems is usually illustrated as a trade-off between translation quality and latency. As in offline translation, BLEU remains the most frequently used metrics for measuring translation quality of online systems. Several metrics have been proposed for latency measurement \cite{cho2016can,Ma19acl,cherry2019thinking}, amongst which Average Lagging (AL) proposed by \cite{Ma19acl} is a frequent choice. The original AL metric measures the average rate at which the translation system lags behind an ideal \textit{wait-0} translator. \cite{simuleval2020} argues that this metric has a shortcoming when applied to $S2T$ translation, and proposes an adaptive version which remedies this problem. However, we noticed that this adaptive version is strongly sensitive to the reference's length, which can be arbitrarily long and weakly dependent on the input speech. In some cases, a slight change of the reference length (which might come from a different tokenization method for example) could drastically change the AL value. Furthermore, one should keep in mind that negative values of AL can still occur when the translation system in question gets ahead of the ideal translator (i.e when it predicts output tokens although the already read source frames do not account for them). Despite those shortcomings, we keep using the adaptive AL from \cite{simuleval2020} in this work in order to measure our improvements of results over those of \cite{nguyen2021empirical_icassp}.

%YE: I suggest to change the title section\section{Extensions of our E2E online model}

\vspace{-10pt}
\section{End-to-end online model}
%\textbf{Model of \cite{nguyen2021empirical_icassp}}
%here describe in a few lines the models proposed in your icassp paper

\textbf{Our previous work \cite{nguyen2021empirical_icassp}} reused an attention-based encoder-decoder architecture described in \cite{nguyen2019ontrac}. The speech encoder stacks two VGG-like CNN blocks \cite{simonyan2014very} before five layers of BLSTM. We stack in each VGG block two 2D-convolution layers, followed by a 2D-maxpooling layer. After these two VGG blocks, the shape $(T \times D)$ of an input speech sequence is transformed to $(T/4 \times D/4)$, with $T$ being the length of the input sequence (number of frames), and $D$ being the features' dimension respectively. The decoder is a stack of two 1024-dimensional LSTM layers, and  Bahdanau's attention mechanism~\cite{bahdanau2014neural} is used to bridge the encoder and the decoder. In online mode, the BLSTM speech encoder %Since their speech encoder uses BLSTM layers, at each decoding step $t$, they needs to re-encode the whole input sequence from the beginning. 
must re-encode from the beginning, from left-to-right and from right-to-left, the input speech sequence every time new input frames are read. In terms of decoding strategy, an adaptive version of \textit{wait-k} is proposed in \cite{nguyen2021empirical_icassp}. This deterministic decoding strategy reads at the first reading operation $k$ (\textit{wait} parameter) first acoustic frames of the input speech features sequence. At each reading operation after this, the system continues consuming fixed intervals of $s$ (\textit{stride} parameter) frames (this reading strategy is also referred in this paper as the fixed interval boundaries segmentation method). A writing operation is put after each reading operation, which writes at maximum $N$ (\textit{write} parameter) output tokens.

%\textbf{[k,s,N should be introduced here !!!!!]}

\begin{figure}[ht]
    \centering
    \begin{subfigure}{.5\textwidth}
    \centering
    \includegraphics[scale=0.5]{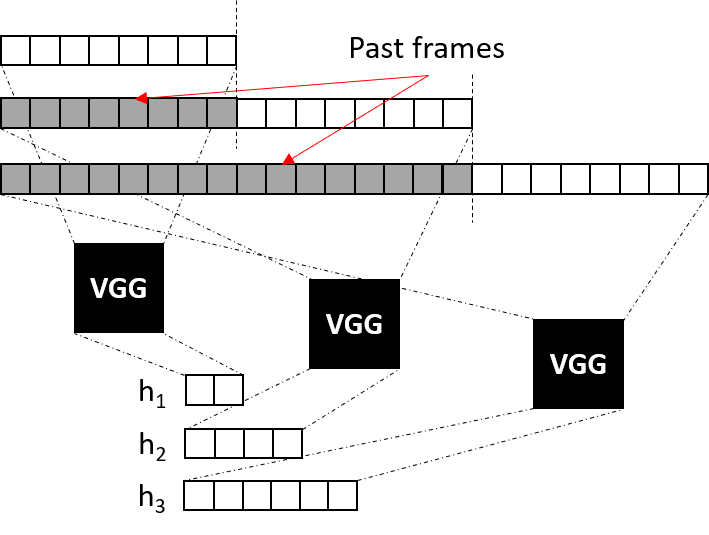}
    \caption{Re-encode}
    \label{fig:reencode}
    \end{subfigure}
    %\begin{subfigure}{.5\textwidth}
    %\centering
    %\includegraphics[scale=0.45]{Figures/read-independently.png}
    %\caption{Read independent chunks of frames}
    %\label{fig:read_independently}
    %\end{subfigure}
    \begin{subfigure}{.5\textwidth}
    \centering
    \includegraphics[scale=0.5]{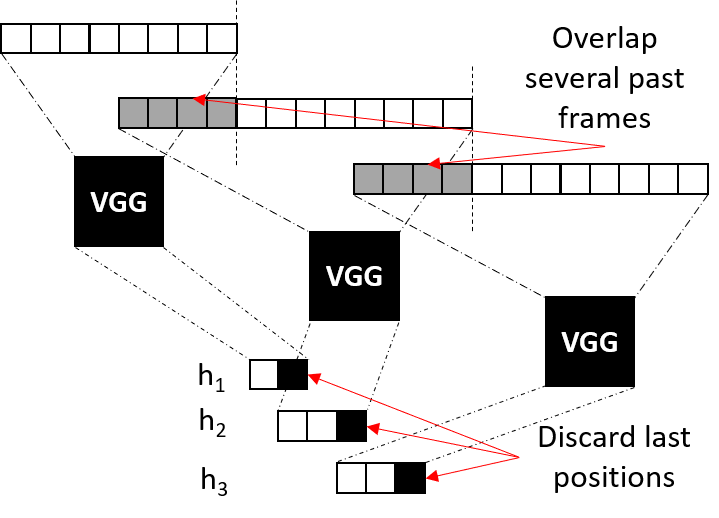}
    \caption{Overlap-and-compensate}
    \label{fig:overlap_and_compensate}
    \end{subfigure}
    \caption{Different encoding strategies.}
    \label{fig:overlap_algorithm}
\end{figure}

\textbf{ULSTM Re-encode strategy}
%\cite{nguyen2021empirical_icassp} investigates the decoding strategy that allows them to leverage a pre-trained offline model whose speech encoder's nature is BLSTM. 
%With a BLSTM speech encoder, \cite{nguyen2021empirical_icassp} must re-encode from the beginning, from left-to-right and from right-to-left, the input speech sequence every time new input frames arrive. However,
\cite{elbayad:hal-02962195} proves that, for $T2T$ online translation, using a ULSTM encoder gives not only better decoding speed but also better BLEU/AL trade-off. We verify if this idea works for speech as well, comparing BLSTM and ULSTM speech encoders in this work. In order to make this comparison, we retrain an offline model similar to the one presented in \cite{nguyen2021empirical_icassp}, except that the speech encoder is modified to stacked ULSTM layers instead of BLSTM layers after the VGG-like blocks. In this strategy (presented in figure \ref{fig:reencode}) we still re-encode the full speech sequence left-to-right every time we read new input frames, but this \textit{ULSTM-Re-encode} approach frees us from computing the BLSTM's right-to-left re-encoding pass, hence being expected to improve decoding speed. 

\textbf{ULSTM Overlap-and-Compensate strategy}
Moving from BLSTM to ULSTM is a first step towards efficiency but re-encoding the full sequence left-to-right each time speech frames are read is still sub-optimal. To avoid this, we tried to feed chunk by chunk of input frames independently %(figure \ref{fig:read_independently}) 
but this solution gave very disappointing results probably because of the quality deterioration of the VGG blocks' output representations due to padding issues near the chunk boundaries (especially in the last several positions of the representations).
%lack , whose output representation directly affects the performance of the LSTM layers, . We suspect that CNN's padding might be responsible for the quality deterioration of the output representation of the VGG blocks, especially at its last several positions. 
Therefore, when dealing with ULSTM speech encoders, we propose an \textit{Overlap-and-Compensate} encoding strategy which allows the encoder to read extra frames from the past in order to compensate some discarded positions in the end of the previous output representation of the VGG-like blocks (figure \ref{fig:overlap_and_compensate}).

\begin{algorithm}[ht]
\SetAlgoLined
\textbf{Input: } sequence $x$; \\
\textbf{Output: } representation $h$; \\
%\KwResult{}
 \textbf{Initialization} step $t=1$, 
 wait parameter $k$, stride parameter $s$, total number of frames read so far $g=k$, $offset=0$,\\
 $finish\_read=False$, $h_0=None$,\\ 
 $overlap=round(k/2)$; \textit{\# Overlap half of chunk\_size}\\
 \While{$g < |x|$}{
   %$overlap=round(s/2)$;$chunk\_size=g-offset$;\\
   %\While{(not $finish\_read$) $\And$ (not OverlapValid($overlap$, $g-offset$))}{

%   \While{(not $finish\_read$) $\And$ ($round(overlap/4) < round(chunk\_size/4)$)}{
%     $t+=1$; $g+=s$; \\
%     $overlap=round(s/2)$;
%   }
  \If{$t>1$}{
    $overlap=round(s/2)$;
   }
  \If{$g>=|x|$}{
    $g = |x|$; $overlap=0$;$finish\_read=True$;
   }
   $x_t=x[offset:g]$;   \textit{\# A chunk read at time $t$}\\
   $h_t=Encode(x_t, overlap, h_{t-1}, finish\_read)$; \\
   $g+=s$; $t+=1$; $offset=g-overlap$;
 }
 
 %\SetKwFunction{FOverlapValid}{OverlapValid}
 %\SetKwProg{Fn}{Fuction}{:}{}
 %\Fn{\FOverlapValid{$overlap$, $chunk\_size$}}{
 %       \textit{\# Divided by $4$ because the encoder has 2 VGG-like blocks} \\
 %       \If{$round(overlap/4) >= round(chunk\_size/4)$}{
 %       \KwRet False;
 %       }
 %       \KwRet True;
 % }
 
 \SetKwFunction{FEncode}{Encode}
 \SetKwProg{Fn}{Fuction}{:}{}
 \Fn{\FEncode{$x$, $overlap$, $prev\_h$, $finish\_read$}}{
    $num\_discard=round(overlap/4)$; \\
    $h_{vgg} = VGG(x)$; \\
    \If{not $finish\_read$}{
      \textit{\# Discard num\_discard positions in the end} \\
      $new\_length=|h_{vgg}|-num\_discard$; \\
      $h_{vgg}=h_{vgg}[0:new\_length]$;
    }
    \KwRet $h_{ULSTM} = ULSTM(h_{vgg}, prev\_h)$;
  }
 \caption{Overlap-and-Compensate encoding strategy}
 \label{algorithm:ulstm_overlap}
\end{algorithm}

Algorithm \ref{algorithm:ulstm_overlap} describes the overlap-and-compensate approach applied to the fixed interval segmentation presented in \cite{nguyen2021empirical_icassp}. It introduces another parameter $overlap$, which decides how many past frames the encoder should read at each encoding step (our \textit{re-encode} strategy corresponds to $overlap=0, offset=0$).
%throughout the whole process). 
We experiment with $overlap$ corresponding to half of the number of input frames of the current step ($overlap=round(s/2)$). %The last VGG block should not throw away all of its representation, therefore, when this happens ($round(overlap/4)>=round(chunk\_size/4)$), we wait for more input frames to come. %(check $OverlapValid()$ function).
%In case the $chunk\_size$ of each step varies (for example the oracle phoneme/word boundaries, or randomly set boundaries segmentation presented in section \ref{sec:experiments}), $s=|segment[t]|-|segment[t-1]|$, and $k=|segment[0]|$. Note that as for the oracle phoneme/word boundaries, $segment[0]$ corresponds to all words read at the first decoding step.

%The comparisons with respect to BLEU/AL trade-off and decoding speed of the strategies using BLSTM, ULSTM re-encode, ULSTM O\&C will be given in the coming sections.

%describe here the extensions over the model described in the ICASSP paper (ULSTM-rencode and ULSTM-overlap)

\vspace{-5pt}
\section{Experimental Setup}
\label{sec:experimental_setup}
\vspace{-5pt}
\textbf{Data} This work focuses on the English-German (EN-DE) language pair. As mentioned in \cite{elbayad:hal-02895893}, the data used to train our models is a combination of MuST-C EN-DE \cite{mustc19}, Europarl EN-DE \cite{europarlst}, and How2 \cite{sanabria18how2} synthetic (i.e. the German translation has been automatically generated by a $T2T$ machine translation system), overall more than 750h of translated speech.

%\textbf{Model architecture:} This work reuses an attention-based encoder-decoder architecture described in \cite{nguyen2019ontrac}. The speech encoder stacks two VGG-like CNN blocks \cite{simonyan2014very} before five layers of LSTM \cite{LSTM}. We stack in each VGG block two 2D-convolution layers, followed by a 2D-maxpooling layer. After these two VGG blocks, the shape $(T \times D)$ of an input speech features is transformed to $(T/4 \times D/4)$, with $T$ being the length of the input sequence (number of frames), and $D$ being the features' dimension respectively. The decoder is a stack of two 1024-dimensional LSTM layers, and  Bahdanau's attention mechanism~\cite{bahdanau2014neural} is used to bridge the encoder and the decoder. 

\textbf{Pre-trained models} The offline BLSTM model presented in this work was trained for our participation to IWSLT 2020 \cite{elbayad:hal-02895893}. It scores $21.38$ and $20.54$ BLEU on MuST-C tst-COMMON, and tst-HE, in greedy decoding mode, respectively. We pre-train another offline ULSTM model with exactly the same configuration as the BLSTM model, only replacing BLSTM layers by ULSTM layers. It scores $18.21$ and $17.98$ BLEU on tst-COMMON, and tst-HE, in greedy decoding mode, respectively.

\vspace{-5pt}
\section{Experiments}
\label{sec:experiments}
%\vspace{-5pt}
\subsection{Impact of encoding strategies}

\begin{figure}[ht]
    \centering
    \begin{subfigure}{.5\textwidth}
    \begin{tikzpicture}
\pgfkeys{/pgf/number format/.cd,1000 sep={}}
\pgfmathsetmacro{\ALWue}{5806}
\pgfmathsetmacro{\Wue}{23.972}

% Wait-infty >  BLEU: 23.972  DAL: 5807  AL: 5807  AP: 1

\begin{axis}[
    height=5.4cm, width=7.5cm, 
    grid=both, y axis line style=-,
    legend style={
        font=\small, 
        legend cell align=left},
    xtick={0,1000,...,4500},
    minor x tick num=1, 
    ytick={2,4,6,8,...,30},
    minor y tick num=1,
    tick label style={font=\small},
    label style={font=\small},
    xmin=0,xmax=4500,
    ymin=0, ymax=22,
    xlabel=Average Lagging (AL) in ms,
    ylabel=BLEU,
    every axis plot/.append style={line width=0.9pt, mark size=2.5pt, mark=square},
    legend to name=s2t
   ]

% Regimes
\foreach \x in {1000, 2000, 3000, 4000}
    \addplot [mark=none, line width=0.5pt, red, forget plot] coordinates {(\x, 0) (\x, 26)};

\addplot [mark=none, black, dashed] coordinates {(-500, 20.54) (5000, 20.54)};
\addlegendentry{offline BLSTM}
\addplot [mark=none, blue, dashed] coordinates {(-500, 17.98) (5000, 17.98)};
\addlegendentry{offline ULSTM}

\addplot[black]
table [y=BLEU,x=AL]{results/tsthe_full_maha_wait_adaptive_AL.dat};
\addlegendentry{$BLSTM$}

\addplot[blue]
table [y=BLEU,x=AL]{results/tsthe_full_maha_ulstm_wait_adaptive_AL.dat};
\addlegendentry{$ULSTM\_Reencode$}

%\addplot[orange]
%table [y=BLEU,x=AL]{results/tsthe_full_maha_ulstm_sbs_wait_adaptive_AL.dat};
%\addlegendentry{$ULSTM\_Overlap$}

%\addplot[green]
%table [y=BLEU,x=AL]{results/tsthe_full_maha_ulstm_finetuned_sbs_half_wait_adaptive_AL.dat};
%\addlegendentry{$ULSTM\_OverlapFT$}

\addplot[red]
table [y=BLEU,x=AL]{results/tsthe_full_maha_ulstm_sbs_half_wait_adaptive_AL.dat};
\addlegendentry{$ULSTM\_Overlap$}

\end{axis}
\node[anchor=south east, scale=.76] at (rel axis cs: 0.9,-0.02) {\pgfplotslegendfromname{s2t}};
%\node[anchor=north west, scale=0.8] at (rel axis cs: 0.15,1.5) {\pgfplotslegendfromname{s2t}};

\end{tikzpicture}
    \caption{MuST-C tst-HE}
    \end{subfigure}
    \begin{subfigure}{.5\textwidth}
    \begin{tikzpicture}
\pgfkeys{/pgf/number format/.cd,1000 sep={}}
\pgfmathsetmacro{\ALWue}{5806}
\pgfmathsetmacro{\Wue}{23.972}

% Wait-infty >  BLEU: 23.972  DAL: 5807  AL: 5807  AP: 1

\begin{axis}[
    height=5.4cm, width=7.5cm, 
    grid=both, y axis line style=-,
    legend style={
        font=\small, 
        legend cell align=left},
    xtick={0,1000,...,4500},
    minor x tick num=1, 
    ytick={2,4,6,...,30},
    minor y tick num=1,
    tick label style={font=\small},
    label style={font=\small},
    xmin=0,xmax=4500,
    ymin=0, ymax=22,
    xlabel=Average Lagging (AL) in ms,
    ylabel=BLEU,
    every axis plot/.append style={line width=0.9pt, mark size=2.5pt, mark=square},
    legend to name=s2t
   ]

% Regimes
\foreach \x in {1000, 2000, 3000, 4000}
    \addplot [mark=none, line width=0.5pt, red, forget plot] coordinates {(\x, 0) (\x, 26)};

\addplot [mark=none, black, dashed] coordinates {(-500, 21.38) (5000, 21.38)};
\addlegendentry{offline BLSTM}
\addplot [mark=none, blue, dashed] coordinates {(-500, 18.21) (5000, 18.21)};
\addlegendentry{offline ULSTM}

\addplot[black]
table [y=BLEU,x=AL]{results/tstcommon_full_maha_wait_adaptive_AL.dat};
\addlegendentry{$BLSTM$}

\addplot[blue]
table [y=BLEU,x=AL]{results/tstcommon_full_maha_ulstm_wait_adaptive_AL.dat};
\addlegendentry{ULSTM\_Reencode}

%\addplot[orange]
%table [y=BLEU,x=AL]{results/tstcommon_full_maha_ulstm_sbs_wait_adaptive_AL.dat};
%\addlegendentry{$ULSTM\_Overlap$}

\addplot[red]
table [y=BLEU,x=AL]{results/tstcommon_full_maha_ulstm_sbs_half_wait_adaptive_AL.dat};
\addlegendentry{$ULSTM\_Overlap$}

\end{axis}
\node[anchor=south east, scale=.76] at (rel axis cs: 0.9,-0.02) {\pgfplotslegendfromname{s2t}};
%\node[anchor=north west, scale=0.8] at (rel axis cs: 0.15,1.5) {\pgfplotslegendfromname{s2t}};

\end{tikzpicture}
    \caption{MuST-C tst-COMMON}
    \end{subfigure}
    \caption{Comparing translation models with BLSTM/ULSTM re-encode/ULSTM Overlap encoding strategies, evaluated on MuST-C tst-HE and tst-COMMON.}
    \label{fig:compare_blstm_ulstm}
\end{figure}
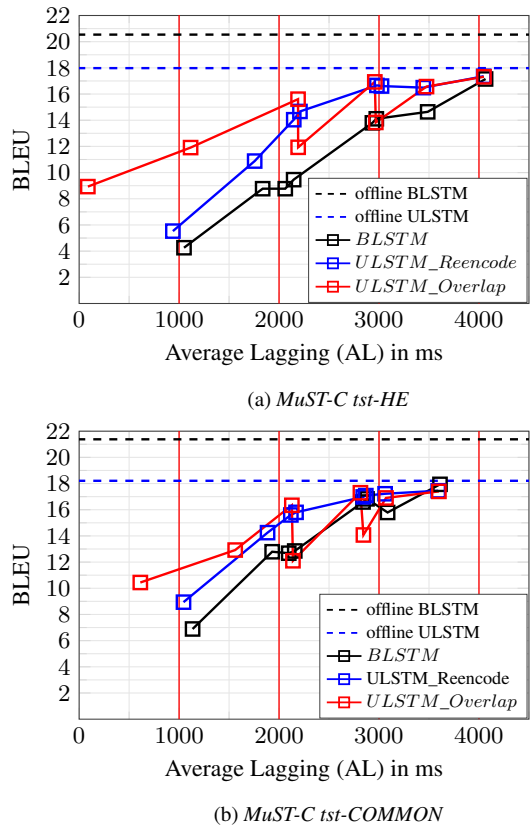

%\begin{figure}[ht]
%    \centering
%    \begin{subfigure}{.5\textwidth}
%    \input{Figures/compare_blstm_ulstm_tst-HE_ter}
%    \caption{MuST-C tst-HE}
%    \end{subfigure}
%    \begin{subfigure}{.5\textwidth}
%    \input{Figures/compare_blstm_ulstm_tst-COMMON_ter}
%    \caption{MuST-C tst-COMMON}
%    \end{subfigure}
%    \caption{Comparing translation models with BLSTM/ULSTM re-encode/ULSTM O\&C encoding strategies, evaluated on MuST-C tst-HE and tst-COMMON. (TER/AL)}
%    \label{fig:compare_blstm_ulstm_ter}
%\end{figure}

This subsection compares different models using either BLSTM or ULSTM speech encoders with different encoding strategies (\textit{re-encode} versus \textit{overlap-and-compensate}). We use the same segmentation (arbitrarly fixed interval boundaries) presented in \cite{nguyen2021empirical_icassp}. 
% which reads at each decoding step (except the first step when a chunk of $k$ frames is read) equal size chunks of $s$ frames in order to generate at maximum $N$ target tokens. 
Figure \ref{fig:compare_blstm_ulstm} illustrates the BLEU/AL trade-off of BLSTM and ULSTM models with different encoding strategies, evaluated on MuST-C tst-HE and MuST-C tst-COMMON, with different $(k, s, N)$ triplets ($k=[100,200]$, $s=[10,20]$, and $N=[1,2]$). It is noticeable that models with ULSTM speech encoders give consistently better BLEU/AL trade-off than the model with BLSTM speech encoder, on both MuST-C tst-HE and tst-COMMON. Moreover, figure \ref{fig:compare_blstm_ulstm} clearly shows that ULSTM \textit{overlap-and-compensate} strategy outperforms ULSTM \textit{re-encode}, especially in low-latency regimes. %(lower than $2000 ms$). %Table \ref{table:overlap_fixed_size} shows that, with the ULSTM \textit{overlap-and-compensate}, using a bigger stride ($s=20$) in order to generate more output tokens ($N=2$) at each step seems more beneficial than using a smaller stride ($s=10$) and generating less output tokens (compare lines 3-4, 5-6 of table \ref{table:overlap_fixed_size}). This explains the few drops witnessed in figure \ref{fig:compare_blstm_ulstm}.

We also investigate the actual time spent decoding each sentence of MuST-C tst-HE using different encoding strategies. In order to do this, we exclusively use the same CPU machine to decode the whole test set using either BLSTM, ULSTM \textit{re-encode} or ULSTM \textit{overlap-and-compensate} encoding strategy. Actual time spent decoding each sentence is captured and averaged over the whole test set. To better illustrate the difference between the encoding strategies, in each latency regime, the time spent of BLSTM is set as a speed unit, and the results of ULSTM \textit{re-encode} and ULSTM \textit{overlap-and-compensate} are reported according to this speed unit.
%Table \ref{table:decoding_speed} shows that, 
We observe that amongst all latency regimes, ULSTM models are much faster than the BLSTM model since they only need to encode the input speech in one direction (from left to right). ULSTM \textit{re-encode} is about twice as fast as BLSTM, scoring $0.53$. Remarkably, scoring $0.06$, the ULSTM \textit{overlap-and-compensate} is fastest among all encoding strategies (about $17$ times faster than the BLSTM, and $9$ times faster than ULSTM \textit{re-encode}, respectively). We believe that this huge improvement in terms of computation speed of the ULSTM \textit{overlap-and-compensate} approach is due to the fact that its input chunks are consistently smaller than that of the ULSTM \textit{re-encode} approach.

%\begin{table}[ht]
%%\centering
%\caption{Decoding speed for models with BLSTM/ULSTM re-encode/ULSTM overlap-and-compensate encoding strategies in different latency regimes, measured by average time spent decoding each sentence of MuST-C tst-HE (millisecond/sentence).}
%\scalebox{0.96}{
%\begin{tabular}[ht]{c|ccc}
%\hline
%\multirow{2}{*}{\textbf{Encoding}} & \multicolumn{3}{c}{\textbf{Latency regime}} \\
%\cline{2-4}
% & \textbf{2000ms} & \textbf{3000ms} & \textbf{4000ms} \\
%\hline
%\hline
%BLSTM & $53140 \pm 213$ & $53187 \pm 67$ & $51314 \pm 37$ \\
%\hline
%ULSTM re-enc & $28028 \pm 70$ & $28136 \pm 184$ & $27140 \pm 77$ \\
%\hline
%ULSTM O\&C & $3323 \pm 40$ & $3337 \pm 21$ & $3216 \pm 55$ \\
%\hline
%\end{tabular}
%}
%\label{table:decoding_speed}
%\end{table}

%\vspace{-5pt}
%\begin{table}[ht]
%%\centering
%\caption{Decoding speed for models with BLSTM/ULSTM re-encode/ULSTM overlap-and-compensate strategies in different latency regimes, measured on MuST-C tst-HE, with the average time spent by BLSTM serves as the time unit, and the average time spent of others are drawn in comparison with this unit.}
%\scalebox{0.88}{
%\begin{tabular}[ht]{|c|c|c|c|}
%\hline
%\multirow{2}{*}{\textbf{Encoding}} & \multicolumn{3}{c|}{\textbf{Latency regime}} \\
%\cline{2-4}
% & \textbf{2000ms} & \textbf{3000ms} & \textbf{4000ms} \\
%%\hline
%\hline
%BLSTM & $1$ & $1$ & $1$ \\
%\hline
%ULSTM re-encode & $0.53$ & $0.53$ & $0.53$ \\
%\hline
%ULSTM overlap-and-compensate & $0.06$ & $0.06$ & $0.06$ \\
%\hline
%\end{tabular}
%}
%\label{table:decoding_speed}
%\end{table}

\vspace{-5pt}
\subsection{Impact of speech input segmentation}

In this section, we investigate the optimal ways to segment the speech flow in order to alternate between reading (R: encoding input) and writing (W: decoding output) operations: fixed interval boundaries (as presented in \cite{nguyen2021empirical_icassp} and in previous experiments of this paper), oracle word boundaries segmentation, and randomly set boundaries segmentation.

%\begin{figure}[ht]
%    \centering
%    \input{Figures/compare_segmentation_ulstm_reencode_tst-HE}
%    \caption{BLEU/AL trade-off of different segmentation methods, evaluated on MuST-C tst-HE, using ULSTM re-encode.}
%    \label{fig:compare_segmentations_ulstm_reencode}
%\end{figure}

%\begin{figure}[ht]
%    \centering
%    \input{Figures/compare_segmentation_ulstm_reencode_tst-HE_ter}
%    \caption{TER/AL trade-off of different segmentation methods, evaluated on MuST-C tst-HE, using ULSTM re-encode.}
%    \label{fig:compare_segmentations_ulstm_reencode_ter}
%\end{figure}

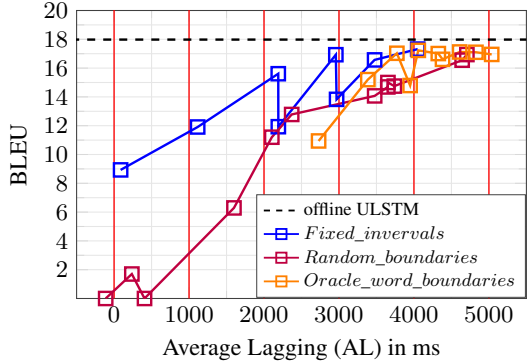
\begin{figure}[ht]
    \centering
    \begin{tikzpicture}
\pgfkeys{/pgf/number format/.cd,1000 sep={}}
\pgfmathsetmacro{\ALWue}{5806}
\pgfmathsetmacro{\Wue}{23.972}

% Wait-infty >  BLEU: 23.972  DAL: 5807  AL: 5807  AP: 1

\begin{axis}[
    height=5.4cm, width=7.5cm, 
    grid=both, y axis line style=-,
    legend style={
        font=\small, 
        legend cell align=left},
    xtick={0,1000,...,5000},
    minor x tick num=1, 
    ytick={2,4,6,8,...,20},
    minor y tick num=1,
    tick label style={font=\small},
    label style={font=\small},
    xmin=-500,xmax=5500,
    ymin=0, ymax=20,
    xlabel=Average Lagging (AL) in ms,
    ylabel=BLEU,
    every axis plot/.append style={line width=0.9pt, mark size=2.5pt, mark=square},
    legend to name=s2t
   ]

% Regimes
\foreach \x in {0,1000, 2000, 3000, 4000, 5000}
    \addplot [mark=none, line width=0.5pt, red, forget plot] coordinates {(\x, 0) (\x, 26)};

\addplot [mark=none, black, dashed] coordinates {(-1000, 17.98) (5500, 17.98)};
\addlegendentry{offline ULSTM}

\addplot[blue]
table [y=BLEU,x=AL]{results/tsthe_full_maha_ulstm_sbs_half_wait_adaptive_AL.dat};
\addlegendentry{$Fixed\_invervals$}
%\addlegendentry{$ULSTM\_Overlap\_EqualSize$}

\addplot[purple]
table [y=BLEU,x=AL]{results/tsthe_full_all_rand_maha_wait_ulstm_sbs_half_adaptive_AL};
\addlegendentry{$Random\_boundaries$}
%\addlegendentry{$ULSTM\_Overlap\_Rand$}

\addplot[orange]
table [y=BLEU,x=AL]{results/tsthe_full_predef_ulstm_sbs_half_maha_wait_adaptive_AL.dat};
\addlegendentry{$Oracle\_word\_boundaries$}
%\addlegendentry{$ULSTM\_Overlap\_Preseg$}

\end{axis}
\node[anchor=south east, scale=.77] at (rel axis cs: 0.93,-0.02) {\pgfplotslegendfromname{s2t}};
%\node[anchor=north west, scale=0.8] at (rel axis cs: 0.15,1.5) {\pgfplotslegendfromname{s2t}};

\end{tikzpicture}
    \caption{BLEU/AL trade-off for different speech input segmentation methods, evaluated on MuST-C tst-HE, using ULSTM \textit{overlap-and-compensate} approach.}
    \label{fig:compare_segmentations_ulstm_overlap}
\end{figure}

%\begin{figure}[ht]
%    \centering
%    \input{Figures/compare_segmentation_ulstm_overlap_half_tst-HE_ter}
%    \caption{TER/AL trade-off of different segmentation methods, evaluated on MuST-C tst-HE, using ULSTM O\&C.}
%    \label{fig:compare_segmentations_ulstm_overlap_ter}
%\end{figure}

%\begin{figure}[ht]
%    \centering
%    \input{Figures/compare_segmentation_blstm_tst-HE}
%    \caption{BLEU/AL trade-off of different segmentation methods, evaluated on MuST-C tst-HE, using BLSTM.}
%    \label{fig:compare_segmentations_blstm}
%\end{figure}

%\begin{figure}[ht]
%    \centering
%    \input{Figures/compare_segmentation_blstm_tst-HE_ter}
%    \caption{TER/AL trade-off of different segmentation methods, evaluated on MuST-C tst-HE, using BLSTM.}
%    \label{fig:compare_segmentations_blstm_ter}
%\end{figure}

\vspace{-5pt}
\subsubsection{Oracle word boundaries}
\vspace{-5pt}
Questioning whether or not feeding relatively precise word-by-word speech chunks instead of fixed-length chunks \cite{nguyen2021empirical_icassp} would improve the performance, we segment the input audio (phrase level) into words using Montreal Forced Aligner \cite{mcauliffe2017montreal}. Their pre-trained English model (from Librispeech \cite{DBLP:conf/icassp/PanayotovCPK15})\footnote{\url{https://montreal-forced-aligner.readthedocs.io}} is used out of the box.
%, without specifying the beam width.
In terms of decoding strategies, we slightly modify the strategy proposed by \cite{nguyen2021empirical_icassp}: 

%\footnote{\url{https://montreal-forced-aligner.readthedocs.io/}}

%\textbf{[LB: reduce what is below...no need to give all those details on failed alignments, etc.]} Their pre-trained English model (from Librispeech \cite{DBLP:conf/icassp/PanayotovCPK15})\footnote{\url{https://montreal-forced-aligner.readthedocs.io/en/latest/pretrained\_models.html#pretrained-acoustic}} is used out of the box, without specifying the beam width. 
%As a result, 588 audio files get aligned successfully with their transcriptions. The remaining 12 audio files cannot be aligned for different reasons, out of which 7 audio files cannot be aligned because the audios lack a word in the end. We removed these words from the corresponding transcription and reran the tool to automatically align these sentences. This left us with 5 audio files, which required manually aligning using Praat\footnote{\url{https://www.fon.hum.uva.nl/praat/}}. 

\begin{itemize}
    \item $k$ remains the number of frames the encoder should wait before starting writing, serving as an upper bound. At the first decoding step, the encoder reads the first several chunks of frames (matching the words boundaries) until $total\_number\_of\_frames \geq k$. In this work, we experiment with $k=[0, 50, 100, 150, 200]$.
    \item $s$ is the number of source words (chunks of frames) read at each decoding step after the first step. In this work, we keep $s=1$ for all our experiments regarding the oracle word boundaries segmentation. %\laurent{[LB: are those phonemes or word boundaries? i think words so then use words everywhere !]}
    \item $N$ remains the maximum number of output tokens (characters) written at each decoding step ($N=[1, 2]$).
\end{itemize}

\subsubsection{Randomly set boundaries}
\vspace{-5pt}
%Exactly opposite to the oracle phoneme/word boundaries segmentation, 
The randomly set boundaries segmentation method cuts the audio input into random sized audio chunks. However, to avoid unreasonable fluctuation of the size of each chunk, we set a lower bound (the minimum number of frames) and a higher bound (the maximum number of frames) for each chunk. The number of frames in each chunk is randomly generated within this constraint. %\textbf{[LB: what follows is unclear, do you first define number of chunks ? if yes how ? if not what is the exact process ?]} 
We continuously accumulate these random numbers until their sum exceeds the total number of frames in the input sequence. The number of frames in the last chunk is adjusted so that the sum of frames in all chunks is equal to the input sequence's length. In this work, we experiment with $[low\_boundary, high\_boundary] = [5,10], [5,20], [5,50],[5,100],[10,50],[10,100]$. We experiment with $N=[1,2]$ in this setup.

Algorithm \ref{algorithm:ulstm_overlap} when applied to these two segmentation methods would slightly change: $s=|segment[t]|-|segment[t-1]|$, and $k=|segment[0]|$. Note that as for the oracle word boundaries, $segment[0]$ corresponds to all words read at the first decoding step.

Figure \ref{fig:compare_segmentations_ulstm_overlap} illustrates that the ULSTM \textit{Overlap-and-compensate} encoding strategy performs best with the fixed interval boundaries segmentation. Surprisingly, the oracle word boundaries segmentation does not seem to be beneficial in comparison with the fixed interval boundaries as it almost always takes bigger AL in order to achieve comparable BLEU scores. We suspect that this happens because the average length of each word ($37$ frames) is much bigger than the stride parameter ($s=10$ or $s=20$ frames) that we set for the fixed interval boundaries. Figure \ref{fig:compare_segmentations_ulstm_overlap} also shows that the randomly set boundaries segmentation perform the worst. Their BLEU scores approach $0$ (the red dots at the bottom of figure \ref{fig:compare_segmentations_ulstm_overlap}) when the segment sizes are too small ($[low\_boundary, high\_boundary] = [5, 10]$).

\vspace{-5pt}
\subsection{Highlighting the most difficult utterances for simultaneous decoding}

%This section analyses the impact of the difficulty of each input sentence on the BLEU/AL trade-off. We use Difficulty Lagging (DL) 
\cite{elbayad2020online} introduced a metric to measure the lagging difficulty of an utterance: after
%sentence complexity. Following \cite{elbayad2020online}, we first 
estimating source-target ($(x,y)$) alignments (for instance with \textit{fast-align} %\laurent{[LB: add a ref]}
\cite{dyer2013simple}), they define a non-decreasing function $z^{align}(t)$, denoting the number of source words needed to translate a target word. This function guarantees that at a given decoding position $t$, $z^{align}(t)$ is larger than or equal to all the source positions aligned with $t$. Lagging difficulty (LD) is then defined as equation (1) below, with $\tau=argmin_t{\{t|z_t=|x|\}}$:
\vspace{-5pt}
\begin{equation}
    LD(x, y) = \frac{1}{\tau}\sum_{t=1}^{\tau}z_t^{align}-\frac{|x|}{|y|}(t-1)
\end{equation}

\begin{figure}[ht]
    \centering
    \begin{tikzpicture}
\pgfkeys{/pgf/number format/.cd,1000 sep={}}
\pgfmathsetmacro{\ALWue}{5806}
\pgfmathsetmacro{\Wue}{23.972}

% Wait-infty >  BLEU: 23.972  DAL: 5807  AL: 5807  AP: 1

\begin{axis}[
    height=5.4cm, width=7.5cm, 
    grid=both, y axis line style=-,
    legend style={
        font=\small, 
        legend cell align=left},
    xtick={-1000, 0,1000,...,6000},
    minor x tick num=1, 
    ytick={2,4,6,8,...,30},
    minor y tick num=1,
    tick label style={font=\small},
    label style={font=\small},
    xmin=-1000,xmax=6000,
    ymin=0, ymax=22,
    xlabel=Average Lagging (AL) in ms,
    ylabel=BLEU,
    every axis plot/.append style={line width=0.9pt, mark size=2.5pt, mark=square},
    legend to name=s2t
   ]

% Regimes
\foreach \x in {-1000 ,0, 1000, 2000, 3000, 4000, 5000, 6000}
    \addplot [mark=none, line width=0.5pt, red, forget plot] coordinates {(\x, 0) (\x, 26)};

\addplot [mark=none, black, dashed] coordinates {(-1000, 17.98) (6000, 17.98)};
\addlegendentry{offline full}
\addplot [mark=none, blue, dashed] coordinates {(-1000, 20.72) (6000, 20.72)};
\addlegendentry{offline 100 easiest}
\addplot [mark=none, orange, dashed] coordinates {(-1000, 14.98) (6000, 14.98)};
\addlegendentry{offline 100 hardest}

\addplot[black]
table [y=BLEU,x=AL]{results/tsthe_full_maha_ulstm_sbs_half_wait_adaptive_AL.dat};
\addlegendentry{$Full$}

\addplot[orange]
table [y=BLEU,x=AL]{results/tsthe_100hardest_maha_wait_adaptive_AL.dat};
\addlegendentry{$100hardest$}

%\addplot[red]
%table [y=BLEU,x=AL]{results/tsthe_20hardest_maha_wait_adaptive_AL.dat};
%\addlegendentry{$20hardest$}

\addplot[blue]
table [y=BLEU,x=AL]{results/tsthe_100easiest_maha_wait_adaptive_AL.dat};
\addlegendentry{$100easiest$}

%\addplot[green]
%table [y=BLEU,x=AL]{results/tsthe_20easiest_maha_wait_adaptive_AL.dat};
%\addlegendentry{$20easiest$}

\end{axis}
\node[anchor=south east, scale=.7] at (rel axis cs: 1.01,-0.02) {\pgfplotslegendfromname{s2t}};
%\node[anchor=north west, scale=0.8] at (rel axis cs: 0.15,1.5) {\pgfplotslegendfromname{s2t}};

\end{tikzpicture}
    \caption{BLEU/AL trade-off scored on different subsets of MuST-C tst-HE based on  Lagging Difficulty (LD).}
    \label{fig:compare_DL}
\end{figure}
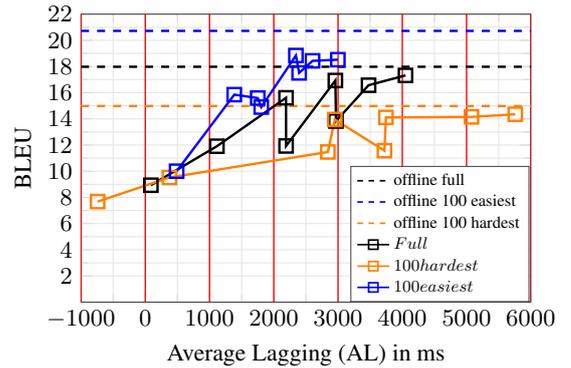

Based on LD, we extract the $100$ most difficult and the $100$ easiest sentences according to the metrics, and report the BLEU/AL trade-off 
%(achieved by the ULSTM O\&C strategy) 
for these sets of utterances.
Figure \ref{fig:compare_DL} shows that LD metrics could be a good tool for highlighting the most difficult utterances for simultaneous decoding since the AL/BLEU curve for the easiest utterances is clearly above the one for the hardest utterances. This suggest the possibility to build specific challenge sets for end-to-end simultaneous speech translation.

%of in general, in the same latency regimes, ULSTM O\&C achieves the best BLEU on 100 easiest sentences, and the worst on 100 hardest sentences.
%\begin{figure}[ht]
%    \centering
%    \includegraphics[scale=0.5]{Figures/t%st_HE_DL_Freq.png}
%    \caption{Frequencies of DLs of tst-HE}
%    \label{fig:tst_HE_DL_freq}
%\end{figure}

\vspace{-5pt}

\vspace{-20pt}
\section{Conclusions}
%\vspace{-5pt}
This paper advocates for using ULSTM instead of BLSTM speech encoder for online translation systems, as it shows that ULSTM outperforms BLSTM in terms of both inference speed and BLEU/AL trade-off. We further improve inference speed and performance of ULSTM speech encoder by proposing a new encoding strategy called ULSTM overlap-and-compensate. Moreover, this work investigates the impact of segmentation on the BLEU/AL trade-off of the ULSTM overlap-and-compensate strategy, and shows that this encoding method works best with equal sized chunks. We also show that difficulty lagging, an indicator of the complexity of the source sentence, might have a great impact on the performance of the online translation systems.
\vspace{-5pt}
\section{Acknowledgements}
%\vspace{-5pt}
This work was funded by the French Research Agency (ANR) through the ON-TRAC project under contract number ANR-18-CE23-0021, and was performed using HPC resources from GENCI-IDRIS (Grant 20XX-AD011011365).

\bibliographystyle{IEEEtran}

\bibliography{mybib}

\end{document}